\documentclass[pmlr]{jmlr}

\usepackage{longtable}
\usepackage{enumitem,kantlipsum}
\usepackage{booktabs}
\usepackage[load-configurations=version-1]{siunitx} 
\usepackage[para]{footmisc}
\usepackage{bm}

\theorembodyfont{\upshape}
\theoremheaderfont{\scshape}
\theorempostheader{:}
\theoremsep{\newline}

\jmlrvolume{1}
\jmlryear{2023}
\jmlrworkshop{VisDA 2022 Challenge: Domain Adaptation for Industrial Waste Sorting}

\title[VisDA 2022]{VisDA 2022 Challenge: Domain Adaptation for \titlebreak Industrial Waste Sorting}

  \author{Dina Bashkirova\nametag{\thanks{Boston University}\thanks{Organizer}},
   Samarth Mishra\footnotemark[1]\footnotemark[2], Diala Lteif\footnotemark[1]\footnotemark[2], Piotr Teterwak\footnotemark[1]\footnotemark[2], \\ Donghyun Kim\footnotemark[1]\footnotemark[2], Fadi Alladkani\nametag{\thanks{Worcester Polytechnic Institute}}\footnotemark[2], James Akl\footnotemark[3]\footnotemark[2], Berk Calli\footnotemark[3]\footnotemark[2], \\ Sarah Adel Bargal\nametag{\thanks{Georgetown University}}\footnotemark[2], Kate Saenko\footnotemark[1]\thanks{MIT-IBM Watson Lab}\footnotemark[2] 
   \\
   Daehan Kim\nametag{\thanks{Hanbat National University}},
   Minseok Seo\nametag{\thanks{SI Analytics}},
   YoungJin Jeon\footnotemark[7],
   Dong-Geol Choi\footnotemark[6]
   \\
   Shahaf Ettedgui\nametag{\thanks{Tel-Aviv University}},
   Raja Giryes\footnotemark[8],
   Shady Abu-Hussein\footnotemark[8] 
   \\
   Binhui Xie\nametag{\thanks{Beijing Institute of Technology}},
   Shuang Li\footnotemark[9]
   }

\begin{document}

\maketitle

\begin{abstract}
Label-efficient and reliable semantic segmentation is essential for many real-life applications, especially for industrial settings with high visual diversity, such as waste sorting. In industrial waste sorting, one of the biggest challenges is the extreme diversity of the input stream depending on factors like the location of the sorting facility, the equipment available in the facility, and the time of year, all of which significantly impact the composition and visual appearance of the waste stream. These changes in the data are called ``visual domains,'' and label-efficient adaptation of models to such domains is needed for successful semantic segmentation of industrial waste. 
 To test the abilities of computer vision models on this task, we present the \textit{VisDA 2022 Challenge on Domain Adaptation for Industrial Waste Sorting}. 
  Our challenge incorporates a fully-annotated waste sorting dataset, ZeroWaste, collected from two real material recovery facilities in different locations and seasons, as well as a novel procedurally generated synthetic waste sorting dataset, SynthWaste. In this competition, we aim to answer two questions: 1) can we leverage domain adaptation techniques to minimize the domain gap? and 2) can synthetic data augmentation improve performance on this task and help adapt to changing data distributions? The results of the competition show that industrial waste detection poses a real domain adaptation problem, that domain generalization techniques such as augmentations, ensembling, etc., improve the overall performance on the unlabeled target domain examples, and that leveraging synthetic data effectively remains an open problem. See
\url{https://ai.bu.edu/visda-2022/}
\end{abstract}
\begin{keywords}
domain adaptation, semantic segmentation, AI for environment
\end{keywords}

\section{Introduction}
\label{sec:intro}

Efficient post-consumer waste recycling is one of the key challenges of modern society as countries struggle to find sustainable solutions to rapidly rising waste levels. World waste production is estimated to reach 2.6 billion tonnes a year in 2030, an increase from its current level of around 2.1 billion tonnes \citep{kaza2018waste}. In the US, one of the leading countries in waste generation by volume, less than 35\% of recyclable waste is being actually recycled \citep{us2017national}, which leads to increased soil and sea pollution and is one of the major concerns of environmental researchers as well as the common public. One of the major challenges in recycling is waste composition analysis and sorting.  In the US and many other countries, recyclable waste is sorted in material recovery facilities (MRFs). MRFs usually use special machinery to automatically sort recyclable waste on a conveyor belt according to the material type, however, they still heavily rely on manual sorting. As such, manual sorting is a mundane, physically demanding, and often dangerous task, as workers are exposed to sharp or contaminated objects on a daily basis. Therefore, an automated solution to aid waste sorting is necessary to make it both safe and profitable, and to ultimately solve the pollution problem. 

\begin{figure}[t] \vspace{-20pt}
    \centering
    \includegraphics[width=1.\linewidth]{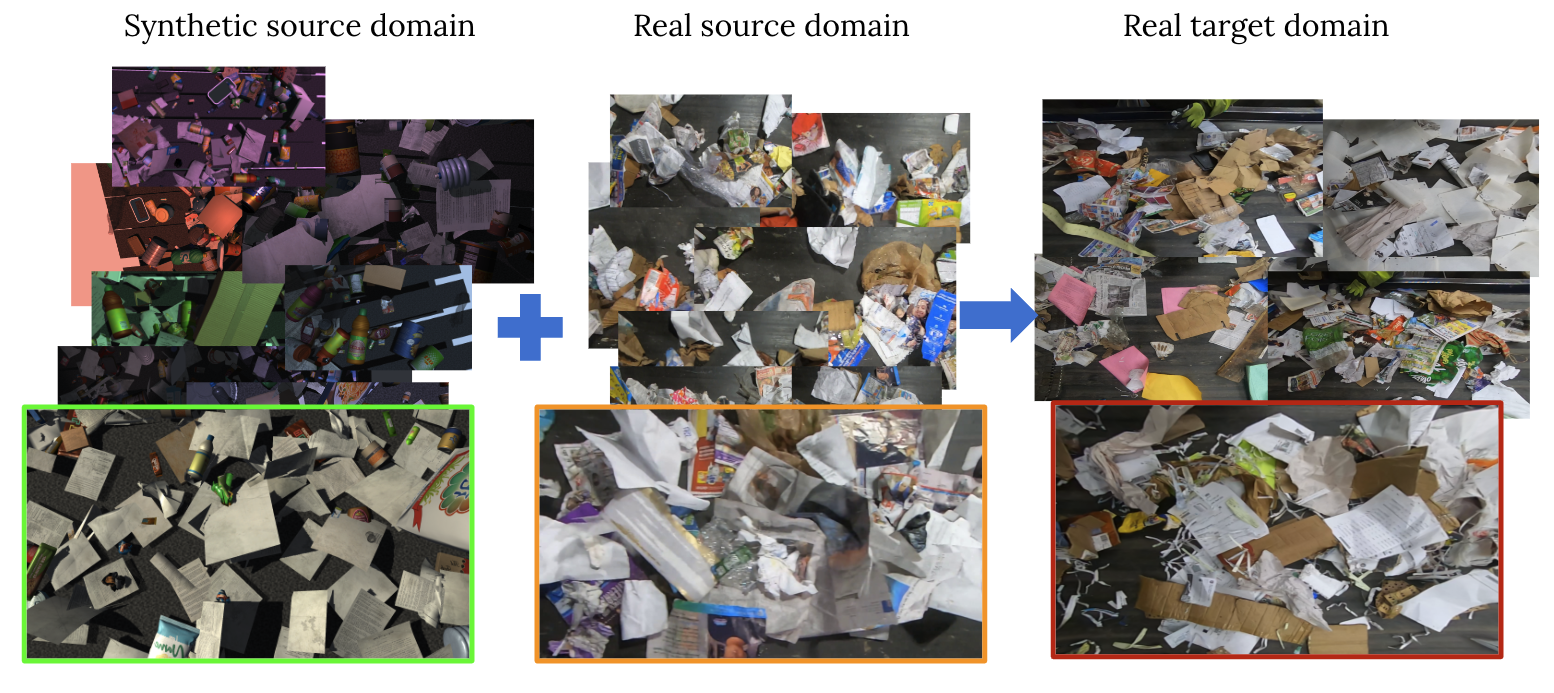}
    \caption{ \vspace{-10pt}
    \textbf{Domain Adaptation for Semantic Segmentation of Recyclables:} Given a large and diverse labeled synthetic dataset \textbf{(left)} and a relatively small labeled real dataset \textbf{(center)} as source domains, the challenge task is to adapt the segmentation model trained on source data to the new unlabeled target domain \textbf{(right)} which introduces a natural domain shift as it was collected at a different location and season than the real source.
    \vspace{-20pt}}
    \label{fig:data_example}
\end{figure}

Computer vision is instrumental for automating waste sorting since it enables segmentation of objects of various material types such as soft plastic, rigid plastic, metal and cardboard. Unfortunately, modern image segmentation models rely on large labeled datasets. It is extremely challenging to collect real in-the-wild waste stream images due to the disruption it causes to the facility's operation. Furthermore, the data annotation for this task involves pixel-level annotation and is prohibitively expensive. 
At the same time, the waste stream varies significantly by object appearance, season, location of the facility, as well as the sorting machinery used at a particular MRF, all of which introduce a significant natural domain shift during deployment and reduce segmentation accuracy. Therefore, \textit{domain adaptation} methods that can adapt a model trained on a labeled \textit{source} dataset to a novel \textit{target} data stream without any additional labels are a promising approach for this problem.

While real data annotation is disruptive and expensive, unlimited amounts of data can be easily generated from 3D models using game engines like Unity, see Figure~\ref{fig:data_example}. Simulation promises to solve the limited and long tailed data problem, but models need to be adapted to an additional visual domain gap, that between non-photorealistic simulations and real images from the MRF. 
 Inspired by the success of simulated training in self-driving applications \citep{richter2016playing, Geiger2012CVPR, cordts2016cityscapes}, we propose a Sim2Real challenge for industrial recycling. The task is to train a segmentation model on a source dataset consisting of a small amount of real data and a large amount of simulated data \emph{and} achieve good results on a held-out real target domain. Algorithms can use the unlabeled target images to improve adaptation. We utilize two fully-labeled datasets for semantic segmentation of recyclable waste: the existing ZeroWaste dataset ~\citep{zerowaste}, a novel ZeroWaste target dataset  (from a different facility and season than the source), and a novel synthetic SynthWaste dataset designed according to the collection protocol of ZeroWaste. We also propose SynthWaste-aug, an augmented version of SynthWaste with instance-level texture augmentations for increased diversity.

\textbf{Relationship to Previous Challenges.}
This challenge is the 6th iteration of the annual VisDA competition. This year, the organizing team consists of researchers from Boston University and Worcester Polytechnic Institute and the challenge was part of NeurIPS 2022 Competitions. A subset of our team also co-organized past VisDA competitions:

\begin{enumerate}[topsep=0pt,itemsep=-1ex,partopsep=1ex,parsep=1ex]
    \item The \href{http://ai.bu.edu/visda-2017/} {1st VISDA (ICCV 2017)} proposed a single-source synthetic-to-real domain adaptation challenge for object classification and semantic segmentation, focusing on  street-view data.
 
    \item The \href{http://ai.bu.edu/visda-2018/} {2nd VISDA (ECCV 2018)} tackled synthetic-to-real open-set domain adaptation for object detection, where the target dataset contained examples of classes that were absent in the source domain.
    \item The \href{http://ai.bu.edu/visda-2019/} {3rd VISDA (ICCV 2019)} introduced  multi-source and  semi-supervised domain adaptation settings of the DomainNet dataset~\citep{peng2019moment} consisting of object classification in six domains (real, clipart, painting, drawing, infograph and sketch).
    \item The \href{http://ai.bu.edu/visda-2020/} {4th VISDA (ECCV 2020)} for domain adaptive instance retrieval, where the target domain had a set of classes (instance IDs) novel with respect to the source domain. 
    \item The \href{http://ai.bu.edu/visda-2021/} {5th VISDA (NeurIPS 2021)} challenge studied a universal domain adaptation setup for object classification, in which the sets of classes in the source and target domains have a significant overlap but both source and target domains have classes that were not present in the other domain. 
\end{enumerate}

Our challenge is different from the previous iterations of VisDA, as it 1)~includes the novel synthetic and real datasets for semantic segmentation of recyclable waste (see Figure~\ref{fig:data_example}), 2)~proposes a setup in which synthetic data is used to improve the  adaptation from one real dataset to another via supervised Sim2Real domain adaptation, as opposed to VisDA 2017 and 2018 that used only synthetic data as a labeled source domain, and 3)~focuses on a challenging application of recyclable waste sorting, a problem that contains different types of distributional shift compared to prior domain adaptation setups, and thus is significantly different from the standard benchmarks. 

Another line of work relevant to the proposed challenge is the simulation-to-real domain adaptation benchmarks, such as GTA~\citep{richter2016playing} or SYNTHIA~\citep{ros2016synthia}-to-KITTI~\citep{Geiger2012CVPR} or Cityscapes~\citep{cordts2016cityscapes}, that focus on the autonomous driving applications. \citep{bousmalis2018using} also proposed to use Sim2Real domain adaptation to improve the quality of robotic grasping. These benchmarks also propose a challenging Sim2Real setup, but on tasks that are different from industrial waste sorting, and therefore, solutions developed for these datasets are not tailored to our task. In addition to that, they propose an unsupervised Sim2Real domain adaptation setup, whereas we aim to leverage limited real supervision to minimize the domain gap and improve generalization to the unseen data in the real domain. 

Another relevant challenge at NeurIPS is the \href{https://www.4paradigm.com/competition/nips2018}{AutoML for Lifelong Machine Learning (NeurIPS'18 competition)} for continuous learning. Although this challenge also addresses the continuously changing data distributions, it is a lifelong AutoML setup that assumes a large-scale labeled dataset similar to the test sets (and in particular, during evaluation, test-set labels were {\em revealed} to the algorithm being evaluated after it made predictions on the most recent batch), whereas our challenge tackles the problem of unsupervised domain adaptation with a large randomized synthetic dataset and a smaller-scale real dataset as source domains.

\section{Challenge Overview}
\subsection{Task}
In this challenge, we propose a Sim+Real domain adaptation task, in which we provide  fully-labeled data from two source domains: the novel large-scale synthetic \textbf{SynthWaste dataset} and a relatively small real \textbf{ZeroWaste dataset} for waste detection. The task at hand is to use these two datasets to adapt the segmentation model to the unlabeled real target domain (\textbf{ZeroWaste-v2}) that introduces a domain shift naturally occurring in the waste sorting application. Models have access to the target data during training. ZeroWaste-v2 is a novel dataset collected according to the ZeroWaste protocol at an MRF at a different location and season. An overview of the proposed task and datasets can be found in Figure \ref{fig:data_example}.

\subsection{Datasets}
Our challenge was based on the following four datasets:
\begin{enumerate}
    \item Real-world \textbf{ZeroWaste} dataset (\href{http://ai.bu.edu/zerowaste/}{http://ai.bu.edu/zerowaste/}) \citep{zerowaste} is an open-access dataset for industrial waste sorting distributed under the \href{https://creativecommons.org/licenses/by/4.0/legalcode}{Creative Commons Attribution 4.0  License}. This dataset consists of $4,503$ fully annotated frames shot at a USA MRF during two hours of its operation. The frames are annotated with polygon semantic segmentation of $4$ classes: cardboard, metal, soft plastic and rigid plastic. All other objects, including paper, as well as the conveyor belt, are labeled as background. 
    \item Synthetic \textbf{SynthWaste} dataset is designed specifically for this challenge to improve generalization and robustness to domain shifts. This dataset consists of $20990$ procedurally generated frames of various recyclable objects randomly spawned onto a conveyor belt using Unity Development Platform that allows free usage for non-commercial purposes. The following simulation parameters are randomized: lighting type, intensity, direction and color, camera angle and position, level of clutter and overall distribution of object classes. 
    
    \item As there are style differences between synthetic and real data, we additionally provide a \textbf{texture-randomized} version of the synthetic \textbf{SynthWaste} dataset (SynthWaste-aug, see Figure~\ref{fig:data_example}). SynthWaste frames are augmented on the instance level using the style transfer-based augmentation with Domain Aware Universal Style Transfer (DAUST)~\citep{hong2021domain}, which further increases visual diversity of waste objects. Objects in a frame are augmented using DAUST using random textures from the Flickr Material Database ~\citep{sharan2009material} according to their material type. 
    
    \item We also collected a real-world \textbf{ZeroWaste-v2} dataset as target domain. This dataset consists of $7,720$ frames collected according to the protocol of ZeroWaste, but in a different season (fall vs spring) and state in the USA (MA vs VT). We annotated $250$ and $1004$ frames for validation and final testing, respectively, and we provide $6,466$ unlabeled frames for training. This novel dataset introduces a real-life domain shift typically occurring in industrial waste sorting. 
    
\end{enumerate}

\subsection{Organization, Metrics and Baselines}
\paragraph{Phases}
The competition consisted of two stages:
\begin{enumerate}[topsep=0pt,itemsep=-1ex,partopsep=1ex,parsep=1ex]
    \item \textbf{Development (June 24 -- September 30):} the labeled training and test sets of ZeroWaste, the SynthWaste and SynthWaste-aug datasets were released to the competitors along with the unlabeled ZeroWaste-v2 training set.
    \item \textbf{Evaluation (September 30th -- October 10th):} the test examples from ZeroWaste-v2 are released, and the teams were asked to submit the prediction results on the unlabeled ZeroWaste-v2 test set to our server, where the solutions are automatically evaluated. 
\end{enumerate}

\paragraph{Metrics and evaluation}



To evaluate the effectiveness of the competing solutions, mean intersection over union (mIoU), the standard semantic segmentation metric, was used to evaluate the performance on the test examples from the \emph{target domain}. We used EvalAI~\citep{yadav2019evalai} for hosting our competition. The mean accuracy of per-pixel predictions (mAcc) was also reported.

\begin{table}[ht]\vspace{-30pt}
    \centering
    \begin{tabular}{l c c c c }
    \toprule
         & train on & eval on  & mIoU  & mAcc  \\ \midrule
     \multicolumn{5}{c}{Source-only Baselines} \\ \midrule
      DeepLabv2 & v1 & v1 & $47.83$ & $60.65$ \\
      DeepLabv2 & v1 & v2 & $30.54$ & $41.72$ \\  
      SegFormer  & v1 & v1 & $56.00$ & $95.45$ \\
      SegFormer  & v1 & v2 & $45.49$ & $91.64$ \\ 
      SegFormer  & Synth+v1 & v2 & $42.61$ & $91.22$ \\ 
      \midrule
      \multicolumn{5}{c}{Adaptation Baselines} \\ \midrule
      DAFormer & v1+v2 & v2 & $\textbf{52.26}$ & $\textbf{91.20}$ \\ 
      DAFormer & Synth+v1+v2 & v2 & $48.31$  & $90.63$ \\
      \midrule
       \multicolumn{5}{c}{Winning Solutions} \\ \midrule
       SI-Analytics (\#1) & v1+v2 & v2 & $\textbf{59.66}$ & $\textbf{92.81}$ \\
       Pros (HRDA) (\#2)  & v1+v2 & v2 & $55.46$ & $92.59$  \\
       BIT-DA(PICO++) (\#3) & v1+v2 & v2 & $54.38$ & $91.80$ \\
      \bottomrule
    \end{tabular}
    \vspace{0.5em}
    \caption{Source-only, baseline domain adaptation results, and the results of the top-3 solutions with ZeroWaste-v1 (\textbf{v1}), ZeroWaste-v2 (\textbf{v2}), and SynthWaste (\textbf{Synth}) datasets. The source-only results of DeepLabv2~\citep{chen2017deeplab} and SegFormer~\citep{xie2021segformer} backbones show that while ZeroWaste-v2 introduces a domain shift that is significant for convnet-based DeepLabv2 architecture, features learned by SegFormer are more robust to this shift. The top submitted solutions are able to improve results significantly above our baselines. \vspace{-20pt}}
    \label{tab:baseline_results_new}
\end{table}

\paragraph{Source-Only Baselines}
We evaluate baseline segmentation models trained only on source data (ZeroWaste-v1 or ZeroWaste-v1+SynthWaste datasets as stated in the second column of Table~\ref{tab:baseline_results_new}) and evaluated on the target data, without any domain adaptation. One such source-only baseline is the transformer-based \textbf{SegFormer}~\citep{xie2021segformer}. Another  is a convolutional network, \textbf{DeepLabv2} \citep{chen2017deeplab}. In Table~\ref{tab:baseline_results_new}, we include the test results on the annotated test set frames from ZeroWaste-v2. Our results indicate that there is a significant domain gap between ZeroWaste-v1 and -v2 when the convnet-based DeepLabv2 is used as a backbone. Notably, the state-of-the-art transformer-based SegFormer is a stronger and more robust to this domain shift, with $10.41\%$ source-only mIoU gap, in contrast to $16.32\%$ gap with DeepLabv2. 
We note that fine-tuning the SegFormer model on SynthWaste slightly improves the overall mean accuracy and obtains similar mIoU as the original SegFormer model pretrained on ImageNet-1K \citep{deng2009imagenet}. We observe that synthetic data improves  performance on frequently occurring classes, such as cardboard and soft plastic.

\paragraph{Domain-Adaptive Baselines}

We used the state-of-the-art \textbf{DAFormer} domain adaptation method by \citep{hoyer2022daformer} trained either on ZeroWaste-v1 or on the combined data consisting of ZeroWaste-v1 and SynthWaste samples, as source domain, and the unlabeled examples from ZeroWaste-v2 as target domain. DAFormer uses the same visual transformer backbone as SegFormer.
It is  evident that the domain adaptation technique introduced in DAFormer improves the mIoU target (v2) domain w.r.t. the SegFormer source-only performance. 

We also see that a naive baseline of training DAFormer on the combined SynthWaste and ZeroWaste-v1 reduces segmentation quality, likely due to a significant domain shift between the real and synthetic datasets. Therefore, the given baselines leave room for improvement, which is what we had hoped to achieve in the proposed challenge. As we will see below, none of the top solutions used the synthetic data, so this remains an open problem.

\paragraph{Materials and code}
We provide a starting kit that includes the data, data loaders and code to reproduce our baseline results at the start of the first phase of the competition.  We also provide the executable used to generate SynthWaste to allow the participants to explore meta-learning approaches to further improve the synthetic data. All the code and materials provided can be found on our github page: \url{https://github.com/dbash/visda2022-org}

\section{Results}
In this section, we provide the main results of our challenge, including the participation statistics, the overview of the winning solutions, as well as the main takeaways. The top three solutions' results are summarized in Table~\ref{tab:baseline_results_new}.

\paragraph{Participation statistics}
$14$ teams actively participated in the development phase of our challenge, with $314$ submissions made total. In the evaluation phase, we limited the total number of submissions per team to avoid overfitting to the test set, and we received $40$ submissions from $8$ competing teams. Based on the results of the evaluation phase, we selected three winning solutions from teams SI-Analytics, Pros, and BIT-DA. 

\paragraph{Reproducibility}
The results of the top-3 solutions according to the evaluation phase were tested and reproduced by the organizing team. The links to the code for reproducing the baseline and top-3 solutions can be found on our website. 

\subsection{First place solution: \texttt{SIA\_Adapt}}

\label{sec:1st}
 The first place solution, \texttt{SIA\_Adapt}, uses DAFormer (the baseline method) as the first step. There were two notable variations to DAFormer used by SIA\_Adapt. First, the team found rare class sampling, which was used in training DAFormer, to be unhelpful for performance, and so they dropped it. Second, and more importantly, instead of an Imagenet-1K pre-trained transformer backbone, an Imagenet-22K pretrained ConvNeXt-L~\citep{liu2022convnet} backbone was used. This allowed the method to use a strong feature representation to build on top of and even without any target data at training time, perform at an impressive 56.4\% mIoU on the target (ZeroWaste v2) test set. As another comparison, when the method was initialized with an Imagenet-1K pre-trained ConvNeXt-L backbone, it achieved 57.29\% mIoU on the target test set, compared to the 59.66\% mIoU (See Table~\ref{tab:baseline_results_new}) achieved with an Imagenet-22K initialization, thus isolating the effect that pre-training had on SIA\_Adapt's performance. To better decouple the effects of the backbone architecture and pre-training, we conducted a study with the DAFormer baseline (See Appendix~\ref{sec:backbone-ablation}).

\begin{figure*}[!t] \vspace{-20pt}
  \centering
  \includegraphics[width=\linewidth]{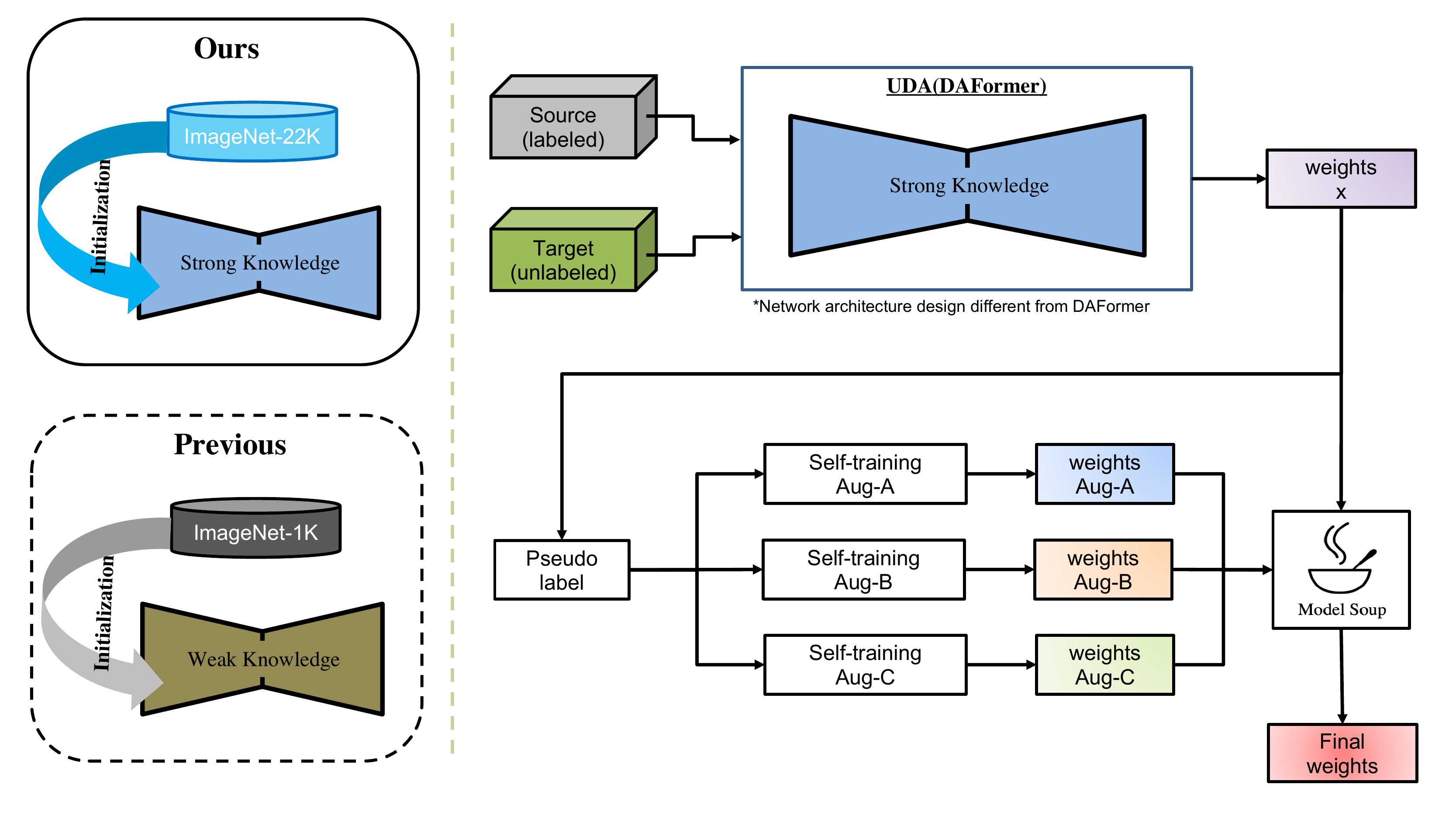}
  \caption{
  \vspace{-3mm}
  Overview of \texttt{SIA\_Adapt}, the first place solution for VisDA-2022. The method uses a strong backbone initialization in the form of an Imagenet-22K pre-trained ConvNeXt-L and pseudolabeling. Also key to the method are self-training using different augmentations and using the resulting models together in a model soup. 
  \vspace{-5mm}
  }
  \label{figure:sia_adapt}
\end{figure*}

With a trained DAFormer (including the modifications as described above), the method proceeds by pseudo-labeling the target and further self-training three different copies of this initial trained DAFormer, each using a different data-augmentation method (See Figure~\ref{figure:sia_adapt}). Finally, these three networks are combined in a model soup~\citep{wortsman2022model}, i.e., their weights are averaged, to obtain the final model. 

\begin{table*}[t] \vspace{-20pt}
    \centering
    \resizebox{\textwidth}{!}{
        \begin{tabular}{ l  c c c c c c }
            \toprule[1.2pt]
            Method & \rotatebox{0}{background} & \rotatebox{0}{rigid plastic} & \rotatebox{0}{cardboard} & \rotatebox{0}{metal} & \rotatebox{0}{soft plastic} & avg. mIoU \\
            \midrule
            Baseline (DAFormer) & $90.78$ & $38.4$ &  $59.73$ & $23.84$ & $48.57$ & $52.26$ \\ \midrule
            SIA\_Adapt & $\bm{92.81}$ & $\bm{48.38}$ & $\bm{65.87}$ & $\bm{36.46}$ & $\bm{54.80}$ & $\bm{59.66}$ \\
            HRDA & $92.20$ & $41.80$ & $63.90$ & $28.30$ & $51.20$ & $55.50$
            \\
            PICO++ & $91.36$ & $44.35$ & $61.71$ & $31.24$ & $43.25$ & $54.38$
            \\
            \bottomrule[1.2pt]
        \end{tabular}
    }
    \caption{Per-category and average mIoU for models trained on ZeroWaste v1 and unlabeled frames from ZeroWaste v2 and evaluated on the ZeroWaste v2 \textbf{test} set. }\vspace{-10pt}
    \label{table:zerov2_class_test}
\end{table*}

\subsection{Second place solution: \texttt{HRDA}}
\label{sec:2nd}
\noindent\textbf{Overview:} HRDA is a context-aware high resolution domain adaptation method for semantic segmentation~\citep{hoyer2022hrda}. The method comprises a multi-resolution training approach for UDA that combines small high-resolution crops and large low-resolution crops to preserve fine segmentation details as well as capture long-range context information. Predictions from both resolution crops are fused using a learned scale attention, which can enable adapting objects at the better-suited scale. As a backbone of their framework, this solution uses the DAFormer~\citep{hoyer2022daformer} architecture that is based on a Transformer network which utilizes self-training. The latter uses pseudo labels generated by a teacher network to iteratively adapt the model to the target domain. Similar to the first place solution, the team concludes after an ablation study in Table \ref{tab:hrda_ab} that rare class sampling (RCS) used to train DAFormer is ineffective for performance.
\\

\noindent\textbf{Results:} Results of this solution are reported in Table \ref{tab:baseline_results_new}, showing that HRDA yields a remarkable improvement in mIoU and mean accuracy compared to the source-only method. In addition, a detailed breakdown of the method's performance is reported in Table \ref{table:zerov2_class_test}. The participating team also provides an ablation study with different source datasets and RCS configurations. In Table \ref{tab:hrda_ab}, the ablation study shows that training on the Zerowaste real-world dataset alone is enough to yield the best performance and that rare class sampling actually deteriorates it.

\begin{table}[h!] \vspace{-10pt}
    \centering
    \begin{tabular}{l c c}
    \toprule
         Source Dataset & RCS & Validation mIoU  \\
         \midrule
         Synthwaste & $\times$ & 20.4\\ 
         Synthwaste + Zerowaste & $\times$ & 45.5 \\
         Synthwaste + Zerowaste, Equal Size & $\times$ & 51.1 \\
         Zerowaste & \checkmark & 47.1 \\
         Zerowaste & $\times$ & \textbf{56.6} \\
         \bottomrule
    \end{tabular}
    \caption{Ablation study by \texttt{HRDA}, the second place solution, examining different source data and the RCS (rare class sampling) configuration of the DAFormer backbone. \vspace{-20pt}}
    \label{tab:hrda_ab}
\end{table}

\subsection{Third place solution: \texttt{PICO++}}

\begin{figure} \vspace{-20pt}
    \centering
    \includegraphics[width=\linewidth]{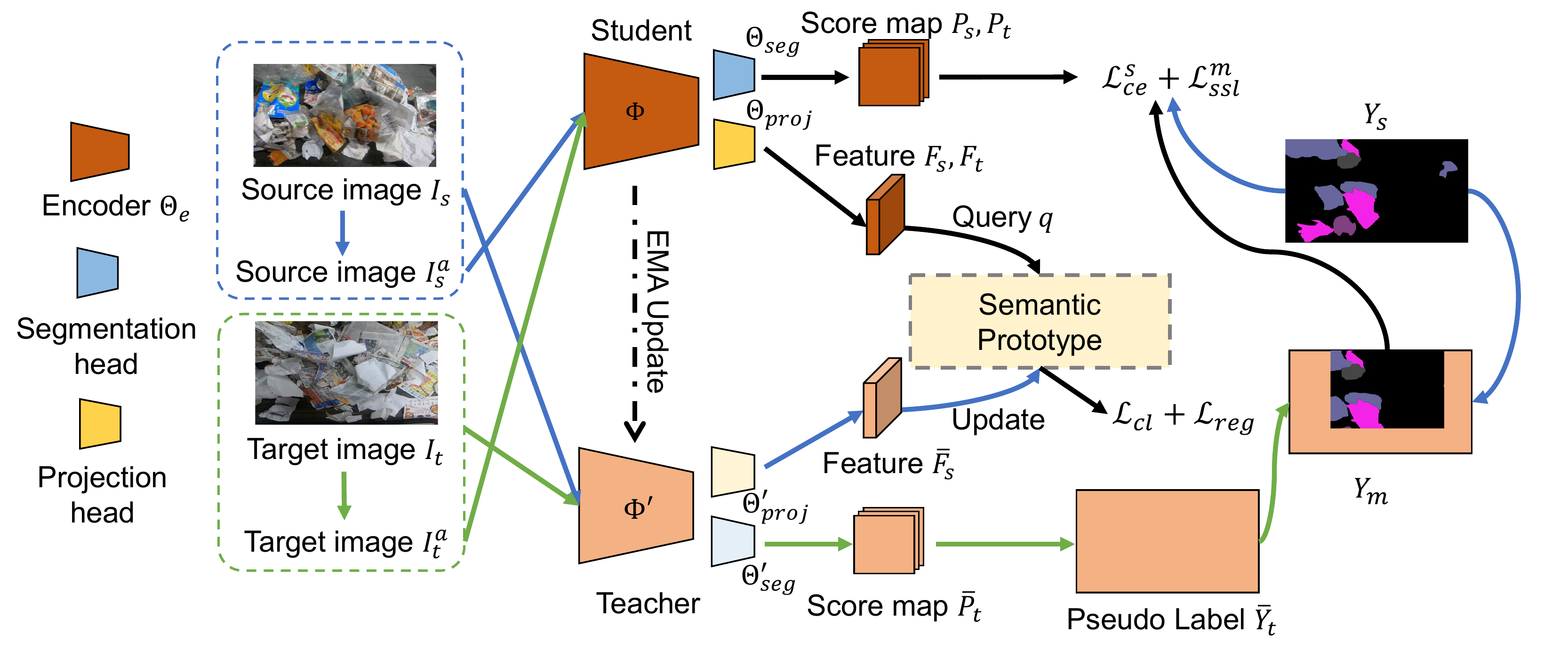}
    \caption{PICO++. A student-teacher framework, where the teacher is updated using EMA. The teacher then is used as a target data pseudo-labeler for both supervised contrastive and cross-entropy losses. Details are in Appendix \ref{sec:app_pico}. \vspace{-10pt}}
    \label{fig:pico++}
\end{figure}

\noindent\textbf{Overview:} PICO++ is a variant of  SePiCo~\citep{xie2023sepico}. The method composes a student-teacher architecture with learning signals from  semi-supervised contrastive and semi-supervised cross-entropy losses.  The student updates the teacher with EMA, while the teacher provides pseudo-labels for the student network to learn from target data. The samples are contrasted against class prototypes, which are computed from teacher representations. Different from other solutions, the contrastive loss provides explicit source-target domain alignment. Similar to other winning solutions,  PICO++ is built on top of the very strong DAFormer~\citep{hoyer2022daformer} architecture. The EMA update provides an implicit ensembling through model parameter averaging, which has been  shown to improve domain generalization performance~\citep{wortsman2021robust}. For details, please see Appendix \ref{apd:solution_details}.
\newline
\newline
\noindent\textbf{Results: }Results of PICO++ are reported in Table \ref{tab:baseline_results_new}. PICO++ achieves substantial performance gains compared to the baseline method. 
As seen in Table \ref{table:zerov2_class_test}, the gain is much greater for classes like rigid plastic and metal, implying that PICO++ has more potential in recognizing classes featuring a relatively regular shape. Nonetheless, a consistent improvement can be observed from all foreground classes, showing the effectiveness of PICO++.
\newline
            

\vspace{-5mm}
\subsection{Lessons learned}
In this challenge, our goal was to investigate which domain adaptation and generalization techniques are particularly efficient for the real-world scenario afforded by the waste detection application. 
Below, we summarize our key observations:

\begin{enumerate}[topsep=0pt,itemsep=-1ex,partopsep=1ex,parsep=1ex]
    \item \textbf{A small real dataset is better than a large synthetic one.} One of the unique features of our challenge compared to the previous Sim2Real competitions was that we propose both the synthetic source domain data and a small-scale real labeled dataset. The main assumption is that SynthWaste has higher visual diversity but is less realistic, and the small-scale ZeroWaste can be used to bridge the gap between the synthetic and real domains. Interestingly, the results from the winning solutions indicate that omitting the sim2real adaptation step and performing domain adaptation from ZeroWaste-v1 to ZeroWaste-v2 results in higher performance. 

    \item \textbf{Pretraining and backbone architecture matter.} The results of the first place solution from SI-Analytics in Sec.~\ref{sec:1st} show a significant boost in segmentation quality when using ImageNet22K-pretrained ConvNeXt-L model compared to an ImageNet1K transformer model used in DAFormer. Our analysis in Appendix~\ref{sec:backbone-ablation} with these changes made in the DAFormer baseline indicate the extent to which each affects target segmentation performance. 
    
    \item \textbf{Ensembling-based techniques and image augmentations are efficient.} The winning solutions commonly use some form of ensembling and / or augmentation, which is shown to improve model generalization. For example, SI-Analytics used model soup \citep{wortsman2022model} of models trained on data augmented with different kinds of augmentations; Pros used HRDA \citep{hoyer2022hrda} that fuses predictions at various resolutions with attention; BIT-DA used a student-teacher paradigm and update the student network with the exponential moving average of the teacher weight update which is an ensembling / regularization technique, and a new variant of a contrastive loss for the source-target domain alignment.
    \item \textbf{DAFormer is a strong baseline.} Even though ZeroWaste-v2 introduces a significant domain shift w.r.t. ZeroWaste with a $17.29\%$ mIoU ($47.83$ versus $30.54$) performance gap with DeepLabv2, DAFormer proved to be a strong baseline, with only $4$ out of $14$ teams beating the baseline in the development phase, and only $3$ out of $8$ in the final evaluation phase. 
    \vspace{-0pt}
\end{enumerate}

\section{Conclusion}
In this paper, we introduced the VisDA 2022 Visual Domain Adaptation Challenge that focuses on domain adaptation for industrial waste sorting. We show that domain shift occurs naturally in the industrial waste detection, and propose a new domain adaptation setup in which a large-scale and diverse synthetic dataset is used alongside the small real dataset to adapt the segmentation model to the real target domain. We propose a novel synthetic dataset, SynthWaste, as well as ZeroWaste-v2 collected according to the protocol of ZeroWaste at a different location and time. Our goal in this challenge was to reach out to the computer vision community to investigate  efficient solutions for pressing and socially important applications and to popularize one of the applications of AI for environment. The results of our challenge suggest that state-of-the-art generalization methods significantly improve the overall performance on the target domain. We believe that our challenge opens new avenues of research in the fields of domain adaptation, and increases awareness and popularity of  environment-centered applications of computer vision.

\paragraph{Acknowledgements}
This work was partially funded by the NSF FW-HTF \#1928477 grant. 
We thank Vitaly Ablavsky for his guidance and support with this project. We are grateful to Nataliia Pyvovar, Guile Domingo and Ahmed Mudawar for their tireless work on data annotation and generation.
\newpage
\bibliography{pmlr-sample}

\appendix

\section{Solution Details}\label{apd:solution_details}

\subsection{SIA\_Adapt}
\label{sec:sia_details}
\paragraph{Datasets}
SIA\_Adapt uses ZeroWastev1 as the labeled source domain and ZeroWastev2 as the unlabeled target domain according to the VisDA 2022 challenge rule\footnote{https://ai.bu.edu/visda-2022/}.
In training, SynthWaste and SynthWaste-aug is not used. 
As per the challenge instructions, ZeroWastev2 test set is used for final evaluation.

\paragraph{Training}
The model was implemented using the DAFormer official code\footnote{https://github.com/lhoyer/DAFormer}\footnote{https://github.com/dbash/visda2022-org}, except for rare class sampling which was not used in SIA\_Adapt.
IN-22K pre-trained weights for ConvNeXt-L are publicly available\footnote{https://github.com/facebookresearch/ConvNeXt}.
An NVIDIA RTX8000 GPU was used for training and all hyperparameter tuning experiments.
40,000 iterations of training was done for the initial adaptive model and 10,000 iterations for the fine-tuned model.

\paragraph{Fine-tuninig}
Model soup recipe was used to combine model weights after self-training. However, no EMA (exponential moving average) was used for training the individual self-trained models.
The 3 different augmentations used, each for a different self-trained model, were: \texttt{PhotoMetricDistortion} implemented by mmseg\footnote{https://github.com/open-mmlab/mmsegmentation}, \texttt{GaussNoise} and \texttt{RandomGridShuffle} implemented by albumentations\footnote{https://github.com/albumentations-team/albumentations}.

\subsection{PICO++}
\label{sec:app_pico}

\noindent\textbf{Student and Teacher Network Architectures:} The student and teacher networks are identical in architecture; they are built on top of the very strong DAFormer architecture, which is also used as a baseline for the challenge.Additionally, an extra projection head is added to reduce dimensionality (512$\rightarrow$256) for both the student and teacher. During training, the student is updated with loss gradients while the teacher is update with an exponential moving average (EMA) of student iterates. 
\newline
\noindent\textbf{Cross-Entropy Losses:} There are two cross-entropy losses used to update the student network. The first is a standard cross entropy loss applied on (augmented) source samples, denoted as $\mathcal{L}_{ce}^s$. Then, augmented target samples are pseudo-labeled using the teacher network, and then mixed with a source sample, creating mixed image $I_m^a$. The target pseudo-label is also mixed with the source label, creating mixed label $Y_m$. Another cross-entropy loss is applied to the student with the resulting mixed image-label pair $\mathcal{L}_{ce}^m(I_m^a, Y_m)$. The ratio of mixed image predictions whose confidence exceed $\beta$, which is called $\lambda_\beta$, reweights the $\mathcal{L}_{ce}^m$. The final loss on mixed images is $\mathcal{L}_{ssl}^m = \lambda_\beta\mathcal{L}_{ce}^m$. This method follows the DACS~\citep{tranheden2021dacs} methodology and allows for self-training using unlabelled target data. 
\newline
\noindent\textbf{Contrastive Losses:} In addition to the cross-entropy losses, a semi-supervised contrastive loss is used following SePiCo \citep{xie2023sepico}. First, the training is warm-started using cross-entropy losses for $T_w = 3000$ iterations. Then, using the source data, per-label gaussians are fit to teacher projection-head features. These per-label gaussians are used to create proto-types to contrast the student features against. For a single sample of class $i$, the contrastive loss is formulated as 
    \begin{align}
        \mathcal{L}_{cl}^{q_i} = -\log\left[\frac{exp(\frac{q_i^\top \mu_i}{\tau} + \frac{q_i^\top \Sigma_i q_i}{2\tau^2})}{exp(\frac{q_i^\top \mu_i}{\tau} + \frac{q_i^\top \Sigma_i q_i}{2\tau^2}) + \sum_{k=1, k\neq i}^{C}exp(\frac{q_i^\top \mu_k}{\tau} + \frac{q_i^\top \Sigma_k q_i}{2\tau^2})}\right] + \frac{q_i^\top\Sigma_iq_i}{2\tau^2}\,,
    \end{align}%
where $q_i$ represents a query feature $q$ of $i^{th}$ class, and $C$ is the class number, $\tau$ denotes the smoothing factor common in contrastive learning. $\mu_i$ and $\Sigma_i$ are the mean and covariance of the distribution within the prototype from $i^{th}$ class. For source samples, ground truth class labels. For target samples, pseudo-labels are used. This contrastive loss ensures cross-domain feature alignment. For the derivation of this loss, please see the SePiCo paper~\citep{xie2023sepico}, section 3.3.2.. Finally the regularization term is formulated as:

    \begin{align}
        \mathcal{L}_{reg}^{\bar{q}} = \frac{1}{C\log C} \sum_{k=1}^{C} \log\frac{e^{\bar{q}^{\top} \mu_k/\tau}}{\sum_{l=1}^{K} e^{\bar{q}^\top \mu_l/\tau}}\,,
    \end{align}%
where $\bar{q}$ is the average of all features for either source or target domain. This prevents collapse of unsupervised samples. 
\newline
\noindent\textbf{Overall loss:} Overall, PICO++ trained with the following objective:

    \begin{align}
        \mathcal{L} = \mathcal{L}_{ce}^s + \mathcal{L}_{ssl}^m + \lambda_{cl}\mathcal{L}_{cl} + \lambda_{reg}\mathcal{L}_{reg}
    \end{align}%
where $\lambda_{cl}$ and $\lambda_{reg}$ are constant weights.

\noindent\textbf{Hyperparameters and training details:}
All  models are trained on a single NVIDIA A100-SXM4-40GB. AdamW~\citep{loshchilov2017decoupled} with betas (0.9, 0.999) and a weight decay of 0.01 is used. The initial learning rate is set to 6e-5 for encoder and 6e-4 for decoder. Note that only the student model is optimized, and the teacher model is momentum updated by the student. DAFormer~\citep{hoyer2022daformer} is followed to employ a learning rate warmup policy in the first 1,500 iterations, and set pseudo confidence threshold $\alpha$ to 0.968, momentum coefficient $\beta$ to 0.999, respectively. The model is trained with a batch of two 640$\times$640-cropped images for 40,000 iterations. To get better initialization of the distributions,  contrastive learning  is started from the 3,000th iteration, and merely update the class prototypes before this. For the weights before loss terms, they are simply fixed to $\lambda_{cl} = \lambda_{reg} = 1$. Similar to the multi-view scheme proposed in DACS~\citep{tranheden2021dacs},  ColorJitter and GaussianBlur is applied only to the samples entering the student model, with a uniform possibility.

\section{Effect of Backbone architecture and Pre-training} \label{sec:backbone-ablation}
Since the first place solution SIA\_Adapt used a strong initialization and architecture in the form of an Imagenet-22K pre-trained ConvNeXt-L backbone, in this section we decouple and isolate the effects of these via experiments using the DAFormer baseline. Table~\ref{tab:app-backbone-ablation} shows the mIoU of DAFormer using different backbone architectures and initializations on the Zerowaste-v2 test set (a 'x' in a column means using the default setting in DAFormer---which corresponds to the MiT transformer backbone and Imagenet-1K pre-training respectively). 

\begin{table}
\begin{center}
\begin{tabular}{c c c}
\toprule
\textbf{ConvNeXt-L backbone} & \textbf{Imagenet-22K pre-training} & mIoU \\
\midrule
 x & x & 52.26 \\
\checkmark & x & 56.41 \\
\checkmark & \checkmark & 58.72 \\
\bottomrule
\end{tabular}
\caption{\label{tab:app-backbone-ablation} Decoupling the effect of backbone architecture and pre-training on Zerowaste-v2 test performance for the DAFormer method.}
\end{center}
\end{table}

\end{document}